\documentclass[10pt,twocolumn,letterpaper]{article}

\usepackage[pagenumbers]{cvpr}   % To produce the CAMERA-READY version (ArXiv建议使用此模式)

\usepackage{etoolbox}
\makeatletter
\@ifundefined{sf@counterlist}{%
  \let\sf@counterlist\@empty
}{}
\makeatother

\usepackage{amsthm}
\usepackage{graphicx}
\usepackage{subfig}

\newtheorem{principle}{Design Principle}

\definecolor{cvprblue}{rgb}{0.21,0.49,0.74}
\usepackage[pagebackref,breaklinks,colorlinks,allcolors=cvprblue]{hyperref}

\title{DisentangleFormer: Spatial-Channel Decoupling for Multi-Channel Vision}

\author{
    Jiashu Liao \thanks{Corresponding author: Jiashu.Liao@glasgow.ac.uk}\\ % 您的名字
    University of Glasgow\\ % 您的第一个机构
    University of Leeds\\   % 您的第二个机构
    {\tt\small Jiashu.Liao@glasgow.ac.uk}\\ {\tt\small J.Liao@leeds.ac.uk}
    \and
    Pietro Liò\\ % 第二位作者名字
    University of Cambridge \\   % 第二位作者的机构
    {\tt\small pl219@cam.ac.uk}
    \and
    Marc de Kamps\\  % 第三位作者名字 (如果有)
    University of Leeds\\   % 第三位作者的机构
    {\tt\small M.deKamps@leeds.ac.uk}
    \and
    Duygu Sarikaya\\  % 第三位作者名字 (如果有)
    University of Leeds\\   % 第三位作者的机构
    {\tt\small D.Sarikaya@leeds.ac.uk}
}

\begin{document}
\maketitle

\begin{abstract}
Vision Transformers face a fundamental limitation: standard self-attention jointly processes spatial and channel dimensions, leading to entangled representations that prevent independent modeling of structural and semantic dependencies. This problem is especially pronounced in hyperspectral imaging, from satellite hyperspectral remote sensing to infrared pathology imaging, where channels capture distinct biophysical or biochemical cues. We propose \textbf{DisentangleFormer}, an architecture that achieves robust multi-channel vision representation through principled spatial–channel decoupling. Motivated by information-theoretic principles of decorrelated representation learning, our parallel design enables independent modeling of structural and semantic cues while minimizing redundancy between spatial and channel streams. Our design integrates three core components: (1) \textbf{Parallel Disentanglement}: Independently processes spatial-token and channel-token streams, enabling decorrelated feature learning across spatial and spectral dimensions, (2) \textbf{Squeezed Token Enhancer}: An adaptive calibration module that dynamically fuses spatial and channel streams, and (3) \textbf{Multi-Scale FFN}:complementing global attention with multi-scale local context to capture fine-grained structural and semantic dependencies. Extensive experiments on hyperspectral benchmarks demonstrate that DisentangleFormer achieves state-of-the-art performance, consistently outperforming existing models on Indian Pine, Pavia University, and Houston, the large-scale BigEarthNet remote sensing dataset, as well as an infrared pathology dataset. Moreover, it retains competitive accuracy on ImageNet while reducing computational cost by 17.8\% in FLOPs. The code will be made publicly available upon acceptance.
\end{abstract}

\begin{figure}[t]
    \centering
    \includegraphics[width=\linewidth]{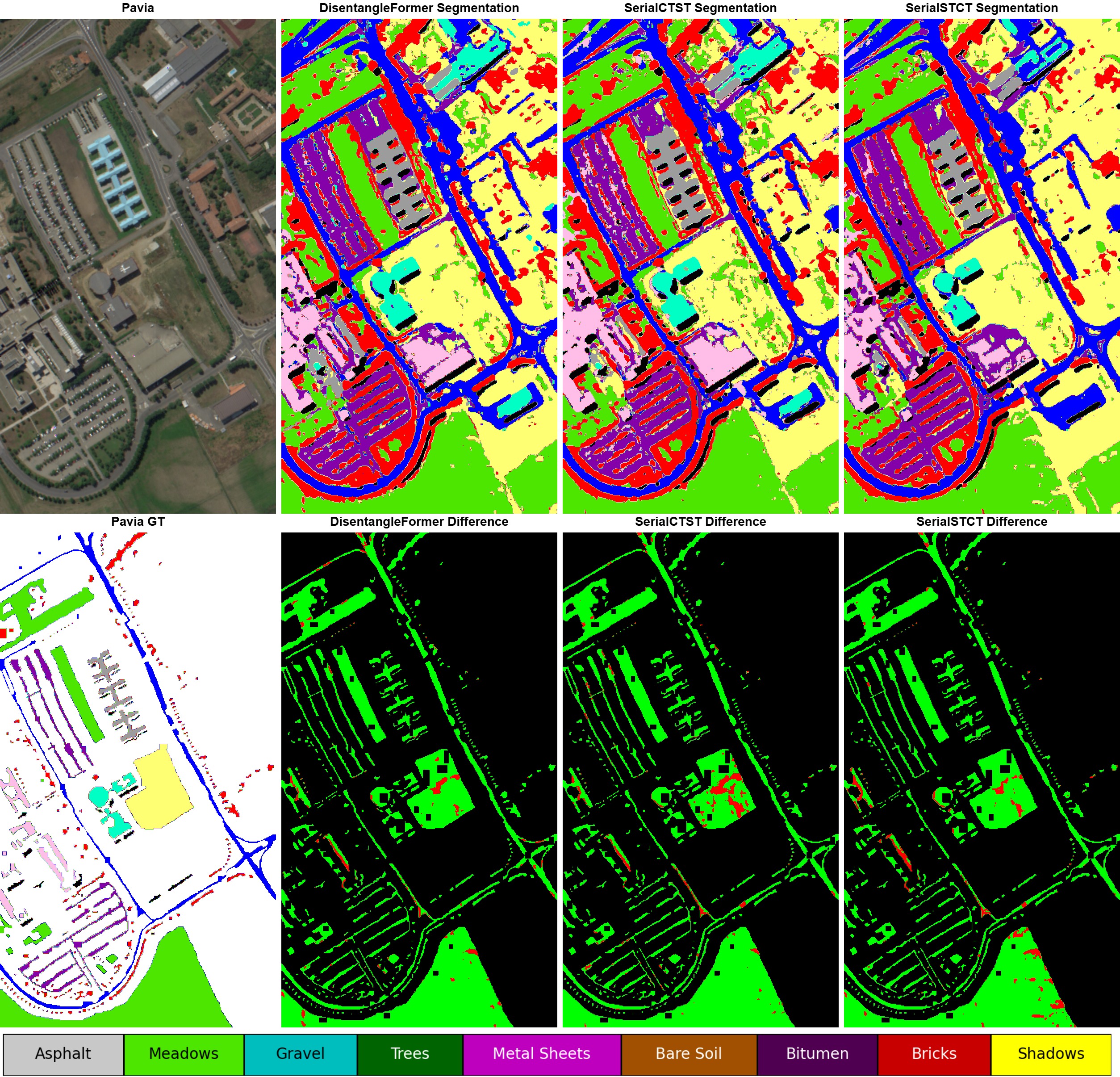}
    \caption{
    Visual validation of Parallel vs. Serial design on Pavia University. Our parallel \textbf{DisentangleFormer (Full)} produces significantly cleaner classification maps with sharper boundaries and less noise compared to the entangled \textbf{SerialCTST} and \textbf{SerialSTCT} baselines.
}
\label{fig:hsi_map_pavia_2}
\end{figure}

\section{Introduction}
\begin{figure*}[ht]
\centering
\includegraphics[width=\textwidth]{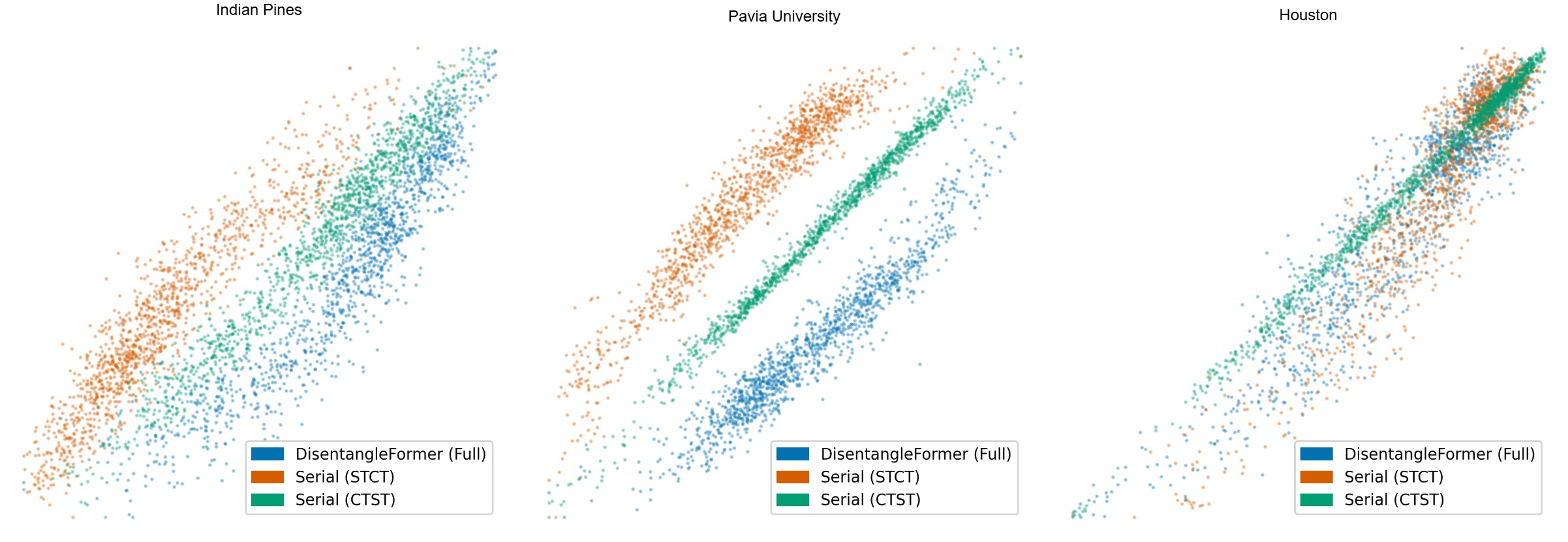}
\caption{
    \textbf{Visual Validation of Information Disentanglement via CCA.}
    This figure compares the first canonical correlation (CCA) scatter plots for DisentangleFormer against the two serial baselines, SerialSTCT and SerialCTST, across all three HSI datasets.
}
\label{fig:cca_composite_plot}
\end{figure*}

The Vision Transformer has transformed modern computer vision, establishing self-attention as a core mechanism for modeling global dependencies. This breakthrough was subsequently advanced by hierarchical Vision Transformers, such as the  widely adopted Swin Transformer~\cite{liu2021Swin}, which introduced local-window attention to mitigate the quadratic complexity of global self-attention while maintaining a multi-scale feature hierarchy. This architectural design has achieved widespread success across computationally intensive domains such as hyperspectral image (HSI) classification~\cite{hong2021SpectralFormer, he2021SpatialSpectral} and large-scale remote sensing (RS) analysis~\cite{papoutsis2022Benchmarking, clasen2025refinedbigearthnet}—both characterized by high-dimensional, multi-channel data. While existing approaches effectively model spatial context (e.g., CNNs, WRNs)~\cite{papoutsis2022Benchmarking, schuhmacher2020CompSegNet, keogan2025Prediction} and spectral sequences (e.g., SpectralFormer, SST)~\cite{hong2021SpectralFormer, he2021SpatialSpectral}, they are constrained by a fundamental architectural limitation: the entangled representation of spatial (structural) and channel (semantic) information within standard attention blocks. By jointly encoding these two distinct data aspects in a single projection space, standard attention forces the model to learn structural and semantic dependencies simultaneously. This entanglement leads to redundant correlations and suboptimal feature utilization, as neither spatial relationships nor channel semantics can be independently optimized. This issue is particularly acute in hyperspectral data domains like satellite or infrared pathology imaging, where individual channels encode distinct biophysical or biochemical signatures.

We argue that effective feature representation requires the decoupling of spatial and channel information streams. Our design framework (Section~\ref{sec:theoretical_foundation}) translates this intuition into concrete architectural principles: representations should preserve informative content within each stream while minimizing redundancy between them. This principle is validated through comprehensive empirical analysis across multiple domains. Motivated by this insight, we introduce \textbf{DisentangleFormer}, a novel Vision Transformer architecture for multi-channel vision through principled information decoupling. Based on this principle, DisentangleFormer instantiates information-theoretic principles via three core components:
\begin{itemize}
    \item \textbf{Parallel Disentanglement Mechanism:} We structurally decouple local window features into two independent processing streams: a Spatial-Token (ST) path and a Channel-Token (CT) path. This parallel design implements our information decoupling principle (Section~\ref{sec:theoretical_foundation}), enabling decorrelated feature learning validated by our ablation studies.
    
    \item \textbf{Adaptive Calibration Module (STE):} To effectively fuse the decoupled streams, we introduce a STE. Structured as a lightweight convolutional network with an adaptive gating mechanism~\cite{woo2018CBAM, hu2018SE}, the STE dynamically recalibrates and merges the ST and CT outputs, enabling more flexible and effective integration than static aggregation techniques.
    
    \item \textbf{Multi-Scale Contextual Feed-Forward Network (MS-FFN):} We replace the standard position-wise MLP with a MS-FFN that employs a multi-branch depthwise convolution operator. This design injects rich, multi-scale spatial inductive biases, enhancing token representations with local contextual information and significantly alleviating the short-range modeling burden on self-attention.
\end{itemize}

\noindent Extensive experiments across a diverse range of multi-channel domains validate the superiority of our decoupled design. DisentangleFormer establishes state-of-the-art (SOTA) performance on three challenging hyperspectral imaging (HSI) benchmarks, the large-scale BigEarthNet remote sensing dataset, and an infrared pathology dataset.

\begin{figure*}[ht]
    \centering
    \includegraphics[width=0.95\textwidth]{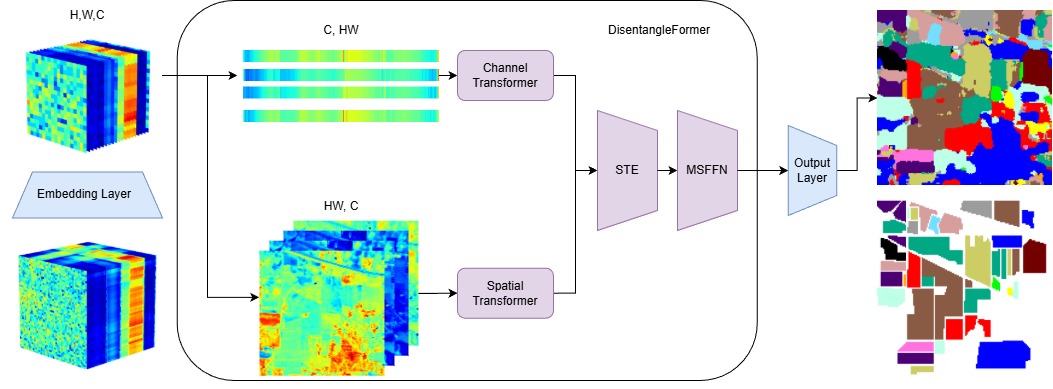}
    \caption{The DisentangleFormer Network Architecture. Input features 
    are processed through an Embedding Layer, then split into parallel 
    Channel Transformer and Spatial Transformer paths. (C, HW) and (HW, C) 
    denote the input dimensions for CT and ST paths respectively. The 
    parallel outputs are fused via the Squeezed Token Enhancer (STE) 
    and processed by the Multi-Scale FFN (MS-FFN). Both transformers 
    employ standard encoder layers with multi-head self-attention. 
    Detailed module structures are provided in the supplementary material.}
    \label{fig:architecture}
\end{figure*}

\section{Related Work}
Our work builds on three research streams: advanced Vision Transformer design, novel attention and fusion mechanisms, and inductive biases for multi-channel data.

\subsection{Architectural Advances in Vision Transformers}
The original Vision Transformer (ViT) introduced self-attention for image representation learning, marking a fundamental shift from convolutional inductive biases to token-based global context modeling \cite{dosovitskiy2020vit}. Due to the computational cost associated with global attention, particularly at high resolutions, subsequent work focused on efficiency. This led to the development of hierarchical architectures like the Swin Transformer \cite{liu2021Swin}, which employs local window-based attention and shift operations for cross-window connectivity, achieving linear complexity with respect to the image size. While effective, the standard Transformer block relies on a monolithic attention operation that tightly couples spatial and channel information, leading to representational redundancy; a core limitation our work addresses. Recent works like Pyramid Vision Transformer (PVT) \cite{wang2021pyramid}, CSWin \cite{dong2022cswin}, and MaxViT \cite{tu2022maxvit} explored various attention patterns but continued to process dimensions jointly. Our approach builds on Swin Transformer’s hierarchical and windowed design for efficiency, but replaces its unified attention block with a disentangled mechanism that explicitly separates spatial and channel feature extraction.

\subsection{Attention Mechanisms and Feature Fusion}
Efficient feature modeling has motivated decoupled attention mechanisms, from channel recalibration (e.g., SE \cite{hu2018SE}) to sequential application (e.g., CBAM \cite{woo2018CBAM}) \cite{wang2020eca, cao2019gcnet, li2019selective, park2018bam}. In the Transformer domain, DaViT architecture \cite{ding2022DaViT} also combines dual attentions, but \textbf{sequentially} ($\mathbf{R}_C = F_{\text{channel}}(F_{\text{spatial}}(\mathbf{X}))$), creating a deterministic dependency. Our architecture differs fundamentally by employing \textbf{parallel independent pathways} ($\mathbf{R}_S = F_{\text{spatial}}(\mathbf{X})$, $\mathbf{R}_C = F_{\text{channel}}(\mathbf{X})$) %directly% 
on the input. This distinction is grounded in our information-theoretic design principles (Section~\ref{sec:theoretical_foundation}) and empirically validated by our ablations (Section~\ref{sec:ablation}, Table~\ref{tab:ablation_studies}B), which confirm parallel processing is superior to sequential entanglement. This theory-driven design, combined with our adaptive STE fusion and MS-FFN, distinguishes our approach from prior dual-attention models.

\subsection{Contextual Modeling through FFNs}
Traditional Transformer FFNs \cite{vaswani2017attention} are position-wise, lacking inherent local spatial context. Recent works remedy this by injecting convolutional inductive bias \cite{liu2022ConvNeXt, Woo2023ConvNeXtV2, graham2021levit}. Building on this, our Multi-Scale FFN (MS-FFN) explicitly addresses this deficit by integrating multi-branch depthwise convolutions \cite{szegedy2015going, chen2017deeplab}, providing rich localized inductive bias to complement the global self-attention mechanism. Unlike hierarchical convolutions (e.g., patch merging) that reduce resolution, our MS-FFN complements self-attention by injecting multi-scale local context in parallel within a single stage.

\subsection{Applications in Multi-Channel Vision}
Our architectural principles are particularly valuable for multi-channel vision tasks where channels encode distinct semantic, biophysical or chemical properties, unlike standard RGB. Such high-dimensional data is common in hyperspectral remote sensing (HSI), multi-modal satellite imagery, and infrared pathology imaging. Hyperspectral Imaging and Remote Sensing: HSI's high spectral resolution requires capturing complex spectral dependencies. While early methods used 3D CNNs or RNNs \cite{hang2019cascaded}, recent Transformers like SST \cite{he2021SpatialSpectral} and SpectralFormer \cite{hong2021SpectralFormer} explicitly model spectral sequences. Large-scale benchmarks like BigEarthNet \cite{papoutsis2022Benchmarking} highlight the need for scalable, specialized architectures. Infrared Pathology: In computational pathology, infrared imaging \cite{keogan2025Prediction, tang2021AdaBoost} provides label-free chemical characterization of tissues. Each spectral channel encodes specific molecular vibrations, creating high-dimensional biochemical fingerprints \cite{raulf2020Deep}. This modality exemplifies our principle: spatial structure (morphology) and spectral signatures (molecular composition) must be modeled independently yet complementarily.

\section{Design Framework and Principles}
\label{sec:theoretical_foundation}
We establish a design framework grounded in information-theoretic principles from disentangled representation learning \cite{higgins2017beta, locatello2019challenging} and mutual information theory \cite{tishby2000information, hjelm2018learning}. 

\textbf{Important Note:} The following principles are \textbf{architectural design guidelines} motivated by information theory, rather than formal mathematical theorems. Their effectiveness is validated through comprehensive empirical analysis (Section~\ref{sec:ablation}, Tables~\ref{tab:ablation_studies}-\ref{tab:cca_validation}, Figures~\ref{fig:hsi_map_pavia_2}-\ref{fig:cca_composite_plot}). Our framework focuses on three key aspects: information decoupling, adaptive fusion, and multi-scale contextualization. See Figure~\ref{fig:architecture} for architecture overview. See detailed model figures in the supplementary material.

\subsection{Information Decoupling Framework}
The conceptual cornerstone of our approach is information decoupling, inspired by disentangled representation learning \cite{higgins2017beta, locatello2019challenging}. We apply principles of decorrelating distinct information sources to supervised multi-channel feature extraction, formalizing that maximizing feature utility requires separating spatial and channel streams while minimizing redundancy. Given a window feature tensor $\mathbf{X} \in \mathbb{R}^{N \times C}$, where $N$ is the number of spatial tokens and $C$ is the channel dimension, we define two fundamental information views: the \textbf{spatial view} $\mathbf{X}_S = \mathbf{X}$ primarily contains structural and positional information, while the \textbf{channel view} $\mathbf{X}_C = \mathbf{X}^T$ primarily contains semantic and feature-correlation information, where the channel view treats spectral bands as ‘tokens’ (C × N) while the spatial view treats spatial patches as tokens (N × C).

\begin{principle}[Information-Guided Spatial-Channel Decoupling]
\label{prin:info_decoupling}
\textbf{Design Principle:} \textit{For effective multi-channel representation, learned spatial and channel features should maximize information content from the input while minimizing redundancy between streams.}
\end{principle}

Formally, given window features $\mathbf{X}$, the ideal representations $\mathbf{R}_S$ (spatial) and $\mathbf{R}_C$ (channel) should satisfy:
\begin{equation}
\begin{aligned}
\text{Maximize:} \quad & I(\mathbf{X}; \mathbf{R}_S) + I(\mathbf{X}; \mathbf{R}_C) \\
\text{while ensuring:} \quad & I(\mathbf{R}_S; \mathbf{R}_C) \leq \epsilon
\end{aligned}
\label{eq:info_objective}
\end{equation}
where $I(\cdot;\cdot)$ denotes mutual information, and $\epsilon$ represents a small redundancy tolerance. The mutual information decomposition $I(\mathbf{X}; \mathbf{R}_S, \mathbf{R}_C) = I(\mathbf{X}; \mathbf{R}_S) + I(\mathbf{X}; \mathbf{R}_C) - I(\mathbf{R}_S; \mathbf{R}_C)$ explicitly balances information preservation with redundancy minimization. 

\textbf{Architectural Implication:} This principle motivates our parallel architecture. Serial processing (e.g., $\mathbf{Y} = F_{\text{spatial}}(F_{\text{channel}}(\mathbf{X}))$) creates deterministic dependency, making $I(\mathbf{R}_S; \mathbf{R}_C) \leq \epsilon$ unattainable. Our parallel design ($\mathbf{R}_S = F_{\text{spatial}}(\mathbf{X})$, $\mathbf{R}_C = F_{\text{channel}}(\mathbf{X})$) enables decorrelated learning by processing both views independently.

\textbf{Empirical Validation:} Our ablation study (Table~\ref{tab:ablation_studies}B) demonstrates that parallel heterogeneous processing consistently outperforms serial alternatives across all datasets. Canonical Correlation Analysis (Table~\ref{tab:cca_validation}, Figure~\ref{fig:cca_composite_plot}) confirms reduced inter-stream correlation, while Figure~\ref{fig:hsi_map_pavia_2} shows qualitative improvements with sharper boundaries.

\subsection{Adaptive Calibration for Decoupled Fusion}
Fusing decorrelated representations effectively requires content-aware mechanisms superior to static fusion functions. Let $\mathbf{R}_S, \mathbf{R}_C$ be the outputs from the spatial and channel pathways. The fused representation $\mathbf{Z} = \text{Concat}(\mathbf{R}_S, \mathbf{R}_C)$ is refined via a dynamic calibration function $\mathcal{F}$: $\mathbf{Y} = \mathbf{Z} + \mathcal{F}(\mathbf{Z})$, where $\mathcal{F}$ generates content-aware adjustments.

\begin{principle}[Adaptive Cross-View Fusion]
\label{prin:adaptive_calibration}
\textbf{Design Principle:} \textit{Fusing decorrelated representations requires content-aware calibration rather than static aggregation to preserve complementary information while suppressing noise.}
\end{principle}

This principle motivates our Squeezed Token Enhancer (STE) module. The STE (Section~\ref{sec:STE}) serves as a lightweight recalibration module that emphasizes salient cross-view interactions through adaptive gating, enabling flexible integration of spatial and channel cues. Table~\ref{tab:ablation_studies}A validates its effectiveness.

\subsection{Multi-Scale Contextualization in FFNs}
Standard Feed-Forward Networks (FFN) in Transformers are position-wise and lack explicit local context modeling. A standard FFN applies a position-wise MLP: $\text{FFN}(\mathbf{y}) = \mathbf{W}_2 \phi(\mathbf{W}_1 \mathbf{y})$. Our Multi-Scale FFN (MS-FFN) operates on the 2D spatial arrangement of tokens $\mathbf{Y}_{\text{2D}} \in \mathbb{R}^{H \times W \times C}$ and employs a multi-branch depthwise convolution operator:

\begin{equation}
\text{MS-DWConv}(\mathbf{Y}') = \bigoplus_{k \in \{1, 3, 5, 7\}} \text{DWConv}_{k \times k}(\mathbf{Y}'_k)
\end{equation}
where $\mathbf{Y}'$ is a projected feature map, and $\oplus$ denotes concatenation. The input channels are split evenly across the parallel branches.

\begin{principle}[Multi-Scale Contextualization]
\label{prin:msffn}
\textbf{Design Principle:} \textit{Explicit multi-scale local context modeling in FFNs reduces the burden on global attention for short-range dependencies, allowing self-attention to focus on long-range relationships.}
\end{principle}

By capturing local spatial correlations at multiple scales, MS-FFN enriches token representations with diverse contextual information. Table~\ref{tab:ablation_studies}A demonstrates significant improvements over standard MLPs.

\noindent These three principles motivate our architecture's core components, validated through comprehensive experiments in Section~\ref{sec:ablation}.

\section{Method}
Our proposed model, which we term \textbf{DisentangleFormer}, builds upon the robust hierarchical architecture of the Swin Transformer. The key innovation lies in replacing the standard self-attention blocks with our newly designed DisentangleFormer blocks, which incorporate a more sophisticated feature extraction and fusion paradigm grounded in the theoretical principles established in Section~\ref{sec:theoretical_foundation}. The model is a 4-stage feature pyramid, stacking DisentangleFormer blocks at each stage and using patch merging layers for downsampling between stages.

\subsection{The DisentangleFormer Block}
The DisentangleFormer block is the core of our model. It processes windowed features $\mathbf{X}_w \in \mathbb{R}^{N \times C}$ (where $N=M \times M$ is the number of tokens in a window) through three main steps: parallel disentanglement, adaptive fusion, and multi-scale feed-forward processing.

\subsubsection{Parallel Feature Disentanglement}
Following Design Principle~\ref{prin:info_decoupling}, we decompose the task into two parallel, structurally independent pathways to model spatial and channel relationships. This component comprises the core of our Disentangled Attention mechanism.

%\begin{itemize} %    \item 

\noindent\textbf{Spatial-Token (ST) Path}: A Transformer Encoder processes the input $\mathbf{X}_w$ directly to capture inter-token spatial relationships.
        \begin{equation}
            \mathbf{R}_S = \text{TransformerEncoder}_s(\text{Norm}(\mathbf{X}_w)) \in \mathbb{R}^{N \times C}
        \end{equation}
%\item 
\textbf{Channel-Token (CT) Path}: Another Transformer Encoder processes the transposed input, $\mathbf{X}_w^T \in \mathbb{R}^{C \times N}$, to model inter-channel correlations. The output is transposed back to align with the spatial path.
        \begin{equation}
            \mathbf{R}_C = \text{TransformerEncoder}_c(\text{Norm}(\mathbf{X}_w^T))^T \in \mathbb{R}^{N \times C}
        \end{equation}
%\end{itemize}

\noindent\textbf{Implementation Details:} Our parallel ST and CT paths 
utilize standard \texttt{nn.TransformerEncoder} layers without 
relative position bias, relying on MS-FFN for local spatial context.

\subsubsection{Adaptive Fusion with STE}
\label{sec:STE}
Following Design Principle~\ref{prin:adaptive_calibration}, the representations from the two disentangled paths are fused using an efficient, content-aware calibration module.

\begin{enumerate}
    \item \textbf{Feature Concatenation}: The spatial and channel representations are reshaped into 2D feature maps and concatenated along the channel dimension.
        \begin{align}
            \mathbf{M}_S, \mathbf{M}_C &= \text{Reshape}(\mathbf{R}_S), \text{Reshape}(\mathbf{R}_C) \in \mathbb{R}^{M \times M \times C} \\
            \mathbf{M}_{\text{fused}} &= \text{Concat}([\mathbf{M}_S, \mathbf{M}_C]) \in \mathbb{R}^{M \times M \times 2C}
        \end{align}
    \item \textbf{Squeezed Token Enhancer (STE)}: The fused map is processed by our STE module, which adaptively recalibrates channel-wise feature responses. The STE is implemented as a lightweight convolutional bottleneck with residual connection:
        \begin{equation}
            \mathbf{M}_{\text{calibrated}} = \text{STE}(\mathbf{M}_{\text{fused}}) \in \mathbb{R}^{M \times M \times 2C}
        \end{equation}
        where STE internally uses depthwise convolution followed by a channel-wise gating mechanism with reduction ratio of 8.
    \item \textbf{Final Projection}: The calibrated map is projected back to the original dimension $C$ and added to the initial input via a residual connection.
        \begin{equation}
             \mathbf{Y}_w = \mathbf{X}_w + \text{Flatten}(\text{Proj}(\mathbf{M}_{\text{calibrated}}))
        \end{equation}
\end{enumerate}

\begin{table*}[ht]
\centering
\caption{Combined performance comparison on all datasets. The arrow ($\uparrow$) indicates that higher values correspond to better performance.}
\label{tab:combined_results}
\begin{tabular}{lrrrrrrrrr}
\toprule
\textbf{Model Name} & \multicolumn{3}{c}{\textbf{Houston}} & \multicolumn{3}{c}{\textbf{Pavia University}} & \multicolumn{3}{c}{\textbf{Indian Pine}} \\
\cmidrule(lr){2-4} \cmidrule(lr){5-7} \cmidrule(lr){8-10}
& \textbf{OA}~$\uparrow$ & \textbf{AA}~$\uparrow$ & \textbf{Kappa}~$\uparrow$ & \textbf{OA}~$\uparrow$ & \textbf{AA}~$\uparrow$ & \textbf{Kappa}~$\uparrow$ & \textbf{OA}~$\uparrow$ & \textbf{AA}~$\uparrow$ & \textbf{Kappa}~$\uparrow$ \\
\midrule
SpectraFormer \cite{hong2021SpectralFormer}  & 0.8801 & 0.8891 & 0.8699 & 0.9107 & 0.9020 & 0.8805  & 0.8176 & 0.8781 & 0.7919 \\
ViT \cite{dosovitskiy2020vit} & 0.7186 & 0.7897 & 0.6804 & 0.7699 & 0.8022 & 0.7010 & 0.8041 & 0.8250 & 0.7876 \\
SwinT \cite{liu2021Swin} & 0.8670 & 0.8904 & 0.8557 & 0.9241 & 0.9300 & 0.8975 & 0.8973 & 0.9375 & 0.8827 \\
A-SPN \cite{9325094} & 0.8840 & 0.8972 & 0.8700 & 0.8387 & 0.7929 & 0.8000 & 0.9356 & 0.8444 & 0.9300 \\
RPNet-RF \cite{s23052499} & - & - & - & 0.9560 & 0.9496 & 0.9427 & 0.9023 & 0.8712 & 0.8887 \\
HyperspectralMAE \cite{Jeong2025HyperspectralMAETH} & - & - & - & - & - & - & 0.9237 & 0.9580 & 0.9100 \\
\textbf{DisentangleFormer (Ours)} & \textbf{0.9262} & \textbf{0.9369} & \textbf{0.9199} & \textbf{0.9651} & \textbf{0.9581} & \textbf{0.9530} & \textbf{0.9611} & \textbf{0.9829} & \textbf{0.9555} \\
\bottomrule
\end{tabular}
\end{table*}

\subsubsection{Multi-Scale Feed-Forward Network (MS-FFN)}
\label{sec:MSFFN}
Following the Disentangled Attention component, we replace the standard MLP-based FFN with our Multi-Scale FFN to enhance local context modeling as specified in Design Principle~\ref{prin:msffn}. The process involves:
\begin{enumerate}
    \item \textbf{Feature Reshaping and Initial Projection}: The output of the attention block, $\mathbf{Y}_{w}$, is reshaped into a 2D feature map $\mathbf{Y}_{\text{2D}} \in \mathbb{R}^{M \times M \times C}$. This map is then projected inwards and activated.
    \begin{equation}
        \mathbf{Z}_{\text{in}} = \phi(\text{Proj}_{\text{in}}(\mathbf{Y}_{\text{2D}}))
    \end{equation}
    where $\phi$ represents activation and normalization.

    \item \textbf{Multi-Scale Convolution with Residual}: The resulting tensor $\mathbf{Z}_{\text{in}}$ is processed by parallel depthwise convolutional paths with kernel sizes $k \in \{1, 3, 5, 7\}$. This multi-scale representation is then added back to the input via a residual connection.
    \begin{equation}
        \mathbf{Z}_{\text{out}} = \phi(\mathbf{Z}_{\text{in}} + \text{MS-DWConv}(\mathbf{Z}_{\text{in}}))
    \end{equation}

    \item \textbf{Final Projection}: The result is projected back outwards to produce the final FFN output.
    \begin{equation}
        \text{MS-FFN}(\mathbf{Y}_{\text{2D}}) = \text{Proj}_{\text{out}}(\mathbf{Z}_{\text{out}})
    \end{equation}
\end{enumerate}
The final output of the DisentangleFormer block is the sum of the FFN output and its input, $\mathbf{Y}_w$.

\section{Experiments}
We validate DisentangleFormer across three demanding domains: HSI classification, large-scale remote sensing, and infrared pathology.

\subsection{Datasets and Evaluation Metrics}
\textbf{High-Spectral Image Classification (HSI):} We evaluate DisentangleFormer on three public, widely-used HSI datasets: Indian Pine (IP)\cite{PURR1947}, Pavia University (PU)\cite{datasetpavia}, and Houston\cite{houston2013}. These datasets are essential for benchmarking spatial–spectral feature extraction given their diverse spatial resolutions and spectral complexity. For all HSI experiments, we report Overall Accuracy (OA), Average Accuracy (AA), and the Kappa Coefficient ($\kappa$). \textbf{Infrared Pathology:} We evaluate our model on molecular infrared (IR) spectroscopy tissue characterization task using the BR20832 breast cancer TMA dataset \cite{tang2021AdaBoost} to differentiate normal and cancerous tissue. We report Accuracy (ACC) and AUC. \textbf{Large-Scale Remote Sensing (RS):} We employ the BigEarthNet dataset \cite{papoutsis2022Benchmarking, clasen2025refinedbigearthnet} for multi-label land-cover classification. This serves as a rigorous test for DisentangleFormer's scalability. We report Average Precision (AP), Mean Average Precision (MAP), Micro F1-Measure ($\mu\text{F\_MF}$), and Micro F1-Measure ($\mu 1$) \cite{papoutsis2022Benchmarking}. \textbf{General Vision Task:} We validate on the widely accepted ImageNet-1K dataset \cite{dosovitskiy2020vit} for general image classification using standard Acc@1 ($\%$) and Acc@5 ($\%$) metrics.

\begin{table*}[h]
\centering
\caption{Image Classification on ImageNet-1K: DisentangleFormer vs. DaViT Baseline. FLOPs measured at $224 \times 224$. Our architecture is specifically optimized for multi-channel imaging tasks, achieving improved computational efficiency (17.8\% FLOPs reduction) while maintaining competitive performance on general vision tasks. The arrow ($\uparrow$) indicates that higher values correspond to better performance. The arrow ($\downarrow$) indicates that lower values correspond to better efficiency.}
\label{tab:imagenet_comparison}
\begin{tabular}{lcccc}
\toprule
\textbf{Model} & \textbf{Acc@1 ($\%$)}~$\uparrow$ & \textbf{Acc@5 ($\%$)}~$\uparrow$ & \textbf{\#Params}~$\downarrow$ & \textbf{FLOPs}~$\downarrow$ \\
\midrule
DaViT-T \cite{ding2022DaViT} & 82.8 & 96.2 & 28.3M & 4.5G \\
DaViT-S \cite{ding2022DaViT} & 84.2 & 96.9 & 49.7M & 8.8G \\
DaViT-B \cite{ding2022DaViT} & \textbf{84.6} & \textbf{96.9} & 87.9M & 15.5G \\
\midrule
\textbf{DisentangleFormer-Tiny (Ours)} & 80.0 & 94.8 & \textbf{25.6M} & \textbf{3.7G} \\
\bottomrule
\end{tabular}
\end{table*}

\begin{table*}[t]
\centering
\caption{Ablation studies on our key DisentangleFormer components. Part A evaluates the necessity of each core component. Part B validates our Design Principle~\ref{prin:info_decoupling} that parallel heterogeneous processing (ST$||$CT) is superior to both serial designs (SerialCTST, SerialSTCT) and parallel homogeneous designs (ST$||$ST, CT$||$CT). Results consistently demonstrate that our full parallel disentangled model achieves optimal performance. The arrow ($\uparrow$) indicates that higher values correspond to better performance.}
\label{tab:ablation_studies}
\resizebox{\textwidth}{!}{%
\begin{tabular}{l|ccc|ccc|ccc}
\toprule
\textbf{Model Variant} & \multicolumn{3}{c|}{\textbf{Houston}} & \multicolumn{3}{c|}{\textbf{Pavia}} & \multicolumn{3}{c}{\textbf{Indian Pine}} \\
\cmidrule(lr){2-4} \cmidrule(lr){5-7} \cmidrule(lr){8-10}
 & \textbf{OA}~$\uparrow$ & \textbf{AA}~$\uparrow$ & \textbf{Kappa}~$\uparrow$ & \textbf{OA}~$\uparrow$ & \textbf{AA}~$\uparrow$ & \textbf{Kappa}~$\uparrow$ & \textbf{OA}~$\uparrow$ & \textbf{AA}~$\uparrow$ & \textbf{Kappa}~$\uparrow$ \\
\midrule
\multicolumn{10}{l}{\textit{Part A: Core Component Ablation}} \\
w/o MS-FFN (use Standard MLP) & 0.9043 & 0.9196 & 0.8961 & 0.9460 & 0.9449 & 0.9272 & 0.9397 & 0.9642 & 0.9309 \\
w/o Channel Path (ST-Only) & 0.9061 & 0.9202 & 0.8981 &  0.9599 & \textbf{0.9632} & 0.9463 &  0.9221 & 0.9686 & 0.9115 \\
w/o Spatial Path (CT-Only) & 0.9115 & 0.9257 & 0.9040 & 0.9489 & 0.9434 & 0.9314 & 0.9447 & 0.9735 & 0.9366 \\
\midrule
\multicolumn{10}{l}{\textit{Part B: Architectural Design Comparison (Serial vs Parallel, Homogeneous vs Heterogeneous)}} \\
\midrule
\multicolumn{10}{l}{\quad \textit{B.1: Serial Heterogeneous Baselines}} \\
\quad Serial: CT $\to$ ST (SerialCTST) &  0.9230 & 0.9345  & 0.9164 & 0.9530 & 0.9440 & 0.9360 & 0.9399 & 0.9487 & 0.9307 \\
\quad Serial: ST $\to$ CT (SerialSTCT) &  0.9205 & 0.9324  & 0.9137 & 0.9532  & 0.9403  & 0.9370 & 0.9453 & 0.9796 & 0.9376 \\
\midrule
\multicolumn{10}{l}{\quad \textit{B.2: Parallel Homogeneous Baselines}} \\
\quad Parallel: ST $||$ ST &  0.8906 & 0.9079  & 0.8812 & 0.9543 & 0.9523 & 0.9499 &  0.9353 & 0.9652 & 0.9260 \\
\quad Parallel: CT $||$ CT &  0.9194 & 0.9315  & 0.9125 & 0.9574 & 0.9536 & 0.9427 & 0.9511 & 0.9715 & 0.9438 \\
\midrule
\textbf{DisentangleFormer (Full: ST$||$CT + STE + MS-FFN)} & \textbf{0.9262} & \textbf{0.9369} & \textbf{0.9199} & \textbf{0.9651} & 0.9581 & \textbf{0.9530} & \textbf{0.9611} & \textbf{0.9829} & \textbf{0.9555} \\
\bottomrule
\end{tabular}
}
\end{table*}

\begin{table*}[h]
\centering
\caption{
    \textbf{Quantitative Validation of Information Disentanglement.} 
    We analyze the first canonical correlation (CCA (1st), $\downarrow$ \textbf{lower means less correlation}) between the ST/CT feature streams of different models on three HSI datasets. DisentangleFormer maintains robustly low CCA scores. In contrast, serial architectures show mixed results: SerialCTST suffers from catastrophic feature entanglement (CCA $\approx$ 1.0), while SerialSTCT achieves the lowest CCA on Houston with the worst performance (see Table~\ref{tab:ablation_studies}).
}
\label{tab:cca_validation}
\resizebox{0.7\textwidth}{!}{
\begin{tabular}{@{}l rrr @{}}
\toprule
{\textbf{Model Architecture}} & \multicolumn{1}{c}{\textbf{Indian Pines}} & \multicolumn{1}{c}{\textbf{Houston}} & \multicolumn{1}{c}{\textbf{Pavia University}} \\ 
& \multicolumn{1}{c}{\textbf{(CCA $\downarrow$)}} & \multicolumn{1}{c}{\textbf{(CCA $\downarrow$)}} & \multicolumn{1}{c}{\textbf{(CCA $\downarrow$)}} \\ 
\midrule
SerialCTST & 0.9776 & 0.9974 & 0.9951 \\
SerialSTCT & 0.9459 & \textbf{0.9302} & 0.9587 \\ 
\midrule
\textbf{DisentangleFormer (Ours)} & \textbf{0.9374} & 0.9325 & \textbf{0.9543} \\
\bottomrule
\end{tabular}
}
\end{table*}

We use \textbf{DisentangleFormer-Tiny} for the ImageNet experiments (Section~\ref{sec:imagenet}). A \textbf{2-stage variant} is used for all other experiments (HSI, Pathology, BigEarthNet). BigEarthNet training follows established protocols \cite{clasen2025refinedbigearthnet}. \textbf{All other architecture and training hyperparameters are detailed in the supplementary material.}

\subsection{Ablation Studies of Architectural Components}
\label{sec:ablation}

We test our architectural design through a comprehensive ablation study (Table~\ref{tab:ablation_studies}), organized into two complementary analyses:

\noindent\textbf{Part A Core Component Ablation:} evaluates each key module. Replacing our Multi-Scale FFN with a standard MLP (w/o MS-FFN) causes consistent performance drops across all benchmarks, validating that offloading local context to FFN improves expressiveness (Design Principle~\ref{prin:msffn}). Removing the channel path (w/o Channel Path (ST-Only)) eliminates inter-channel correlation modeling, particularly detrimental for HSI tasks where spectral relationships are critical. Similarly, removing the spatial path (w/o Spatial Path (CT-Only)) degrades spatial structure understanding, though less severely than removing the channel path for spectral imaging.

\noindent\textbf{Part B Architectural Design Comparison:} validates Design Principle~\ref{prin:info_decoupling} through controlled experiments. First, Serial Heterogeneous Baselines (B.1) like SerialCTST (CT$\rightarrow$ST) and SerialSTCT (ST$\rightarrow$CT) significantly underperform our parallel design across all datasets. This confirms that serial architectures create deterministic feature entanglement ($\mathbf{R}_S = f(\mathbf{R}_C)$ or vice versa), making the redundancy minimization constraint $I(\mathbf{R}_S; \mathbf{R}_C) \leq \epsilon$ unattainable. Second, Parallel Homogeneous Baselines (B.2), including ST$||$ST and CT$||$CT, underperform our heterogeneous design. ST$||$ST performs worst as identical spatial paths learn redundant representations. This confirms true disentanglement requires processing fundamentally different views (spatial vs channel), not merely parallel processing of the same view.The ablation study confirms: (1) all three components are necessary, (2) parallel processing is superior to serial for decorrelated representations, and (3) heterogeneous paths (ST$||$CT) are essential and homogeneous paths (ST$||$ST or CT$||$CT) are insufficient.

\begin{table}[htp]
\centering
\caption{BigEarthNet Multi-label Classification Performance \cite{clasen2025refinedbigearthnet}. The arrow ($\uparrow$) indicates that higher values correspond to better performance.}
\label{tab:bigearthnet_full}
\resizebox{\columnwidth}{!}{
\begin{tabular}{lcccc}
\toprule
\textbf{Model} & \textbf{AP}~$\uparrow$ & \textbf{MAP}~$\uparrow$ & \textbf{$\mu$F\_MF}~$\uparrow$ & \textbf{$\mu 1$}~$\uparrow$ \\
\midrule
ResNet-50 \cite{he2016resnet} & 70.72 & 85.86 & 64.74 & 76.34 \\
ResNet-101 \cite{he2016resnet} & 70.63 & 85.92 & 64.19 & 76.13 \\
MLP-Mixer-Base \cite{tolstikhin2021MLPMixer} & 67.77 & 84.32 & 62.49 & 74.59 \\
MobileViT-S \cite{Qin2024MobileNetV4U} & 69.84 & 86.2 & 62.1 & 75.99 \\
\textbf{DisentangleFormer} & \textbf{71.51} & \textbf{87.25} & \textbf{65.73} & \textbf{77.64} \\
\bottomrule
\end{tabular}
}
\end{table}

\subsection{Main Results on HSI Classification}
\label{sec:main_results}
We compare DisentangleFormer against strong baseline models, including the Vision Transformer (ViT) \cite{dosovitskiy2020vit} and SpectralFormer \cite{hong2021SpectralFormer}, as well as the published State-of-the-Art results across the three benchmarks.

Table~\ref{tab:combined_results} summarizes the performance results on three representative HSI benchmarks. DisentangleFormer consistently achieves state-of-the-art performance across all datasets and metrics, confirming the effectiveness of our principled feature disentanglement approach for multi-channel data. DisentangleFormer achieves OA 0.9262 and Kappa 0.9199, surpassing A-SPN (OA 0.8840, Kappa 0.8700) on Houston. On Pavia University, DisentangleFormer achieves OA 0.9651, outperforming HyperspectralMAE (OA 0.9560) \cite{Jeong2025HyperspectralMAETH}. On Indian Pine, DisentangleFormer gets OA 0.9611, showcasing substantial gains over A-SPN (OA 0.9356) and HyperspectralMAE (OA 0.9237), confirming the architectural innovations provide necessary robustness for real-world HSI challenges.

To visually validate our core claim (Design Principle~\ref{prin:info_decoupling}) and the ablation results from Table~\ref{tab:ablation_studies}, we present a targeted qualitative comparison in Figure~\ref{fig:hsi_map_pavia_2} (See more qualitative results in the supplementary material). We compare our parallel \textbf{DisentangleFormer} against the two serial (entangled) baselines on the challenging Pavia University dataset. The visual results strongly corroborate our quantitative findings: our parallel design produces significantly sharper class boundaries and more accurate spatial details compared to the serial models, which exhibit more classification noise. Full qualitative comparisons against reproduced baseline models for all three HSI datasets (Indian Pine, Pavia University, and Houston) are provided in the supplementary material.

We validate our Design Principle~\ref{prin:info_decoupling} by measuring the information redundancy between the ST and CT feature streams using Canonical Correlation Analysis (CCA), which quantifies the peak linear correlation achievable between two high-dimensional spaces by finding their most correlated linear projections (the first canonical variables). As shown in Table~\ref{tab:cca_validation}, this analysis reveals a fundamental architectural limitation of serial processing. The SerialCTST architecture inherently suffers from deterministic feature entanglement. This is visually confirmed in Figure~\ref{fig:cca_composite_plot} by its collapsed linear distribution. While the SerialSTCT configuration achieves the lowest CCA on Houston (0.9302), this corresponds to the worst classification accuracy (reported in our ablation, Table~\ref{tab:ablation_studies}). This "ineffective disentanglement" demonstrates that minimizing correlation alone is a sub-optimal objective. In contrast, DisentangleFormer is the only architecture that robustly finds an optimal balance, consistently achieving low CCA scores that correlate with its superior classification performance. 

\begin{table}[htp]
\centering
\caption{Infrared Pathology (BR20832 \cite{tang2021AdaBoost}) Performance. The arrow ($\uparrow$) indicates that higher values correspond to better performance.}
\label{tab:pathology_full}
\begin{tabular}{lrr}
\toprule
\textbf{Model Name} & \textbf{ACC ($\%$)}~$\uparrow$ & \textbf{AUC}~$\uparrow$ \\
\midrule
ResNet18 & 88.25 & 0.9520 \\
SwinT \cite{liu2021Swin} & 90.75 & 0.9686 \\
\textbf{DisentangleFormer (Ours)} & \textbf{94.94} & \textbf{0.9865} \\
\bottomrule
\end{tabular}
\end{table}

\subsection{Remote Sensing and Infrared Pathology}
\textbf{BigEarthNet Scalability:} Table~\ref{tab:bigearthnet_full} demonstrates DisentangleFormer's robustness in the large-scale, multi-label domain. The DisentangleFormer model achieves a MAP of 87.25 and a $\mu 1$ of 77.64, surpassing optimized ResNet-101 (MAP: 85.92, $\mu 1$: 76.13) \cite{he2016resnet}. These results establish DisentangleFormer as a new SOTA solution for this large-scale, multi-label benchmark.

\noindent\textbf{Infrared Pathology:} (Table~\ref{tab:pathology_full}) shows the effectiveness of our principle on molecular IR spectroscopy (BR20832) \cite{tang2021AdaBoost}. DisentangleFormer achieves state-of-the-art results with ACC 94.94\% and AUC 0.9865. DisentangleFormer significantly outperforms SwinT (ACC: 90.75\%) \cite{liu2021Swin}. This validates that spatial-channel disentanglement captures subtle molecular signatures in infrared pathology imaging.

\noindent\textbf{Ethics Statement:} The BR20832 dataset used in our pathology experiments is publicly available and has been ethically approved by the original study.

\subsection{General Vision Domain and Potential Limitations}
\label{sec:imagenet}
We analyze the efficiency and generalization capability of the DisentangleFormer architecture as a vision backbone using the ImageNet-1K benchmark (Table~\ref{tab:imagenet_comparison}). While our architecture is primarily optimized for multi-channel, high-dimensional imaging tasks (HSI, remote sensing, infrared pathology), this evaluation serves to validate its computational efficiency and general-purpose capability.

As shown in Table~\ref{tab:imagenet_comparison}, DisentangleFormer-Tiny achieves 80.0\% Acc@1 and 94.8\% Acc@5. While the Acc@1 is lower than the DaViT-T baseline (82.8\%), \textbf{it is important to note that our result was obtained from a single training run without hyperparameter tuning} due to computational limitations. Nonetheless, our model attains this \textbf{competitive accuracy} with dramatically reduced computational cost: it uses only 3.7G FLOPs, representing a \textbf{17.8\% reduction in computational overhead} compared to DaViT-Tiny (4.5G FLOPs), and decreases the parameter count to 25.6M (vs. 28.3M). This result validates that our architectural design is highly efficient and capable of generalizing beyond its primary specialization. This deliberate focus on multi-channel data yields state-of-the-art benefits in our target domains, while its full potential on general vision tasks remains a promising area for future exploration.

\section{Conclusion}
We proposed DisentangleFormer, which adopts principled spatial-channel decoupling guided by information-theoretic design principles. The architecture aims to preserve information content within each stream while minimizing redundancy through parallel independent pathways. DisentangleFormer demonstrated consistent improvements across a wide spectrum of multi-channel vision tasks. It established new state-of-the-art results across all targeted domains: on key HSI classification benchmarks, on the large-scale BigEarthNet remote sensing dataset, and on the specialized BR20832 infrared pathology task. These results demonstrate that this principle-guided design yields robust and efficient models for multi-channel vision tasks. This efficiency was confirmed on ImageNet, where the model reduced FLOPs by 17.8\% compared to DaViT-Tiny while maintaining competitive accuracy. As the design is highly tailored for multi-channel data, its broader performance on general vision tasks remains underexplored, marking a clear avenue for future investigation.

\section*{Acknowledgments}
This work was funded from an EPSRC Healthcare Technologies Network Plus Grant: ``Integrating Clinical Infrared and Raman Spectroscopy with digital pathology and AI: CLIRPath-AI'' (EP/W00058X/1), pump priming award.

{
    \small
    \bibliographystyle{ieeenat_fullname}
    \bibliography{main}

@String(CVPR= {IEEE Conf. Comput. Vis. Pattern Recog.})

@String(ICCV= {Int. Conf. Comput. Vis.})

@String(ECCV= {Eur. Conf. Comput. Vis.})

@String(ICLR = {Int. Conf. Learn. Represent.})

@String(CVPR  = {CVPR})

@String(ICCV  = {ICCV})

@String(ECCV  = {ECCV})

@String(ICLR  = {ICLR})

@article{Jeong2025HyperspectralMAETH,
  title={HyperspectralMAE: The Hyperspectral Imagery Classification Model using Fourier-Encoded Dual-Branch Masked Autoencoder},
  author={Wooyoung Jeong and Hyun Jae Park and Seonghun Jeong and Jong Wook Jang and Tae Hoon Lim and Dae Seoung Kim},
  journal={ArXiv},
  year={2025},
  volume={abs/2505.05710},
  url={https://api.semanticscholar.org/CorpusID:278481063}
}

@inproceedings{dosovitskiy2020vit,
  title={An Image is Worth 16x16 Words: Transformers for Image Recognition at Scale},
  author={Dosovitskiy, Alexey and Beyer, Lucas and Kolesnikov, Alexander and Weissenborn, Dirk and Zhai, Xiaohua and Unterthiner, Thomas and  Dehghani, Mostafa and Minderer, Matthias and Heigold, Georg and Gelly, Sylvain and Uszkoreit, Jakob and Houlsby, Neil},
  booktitle={ICLR},
  year={2021}
}

@inproceedings{liu2021Swin,
  title={Swin Transformer: Hierarchical Vision Transformer using Shifted Windows},
  author={Liu, Ze and Lin, Yutong and Cao, Yue and Hu, Han and Wei, Yixuan and Zhang, Zheng and Lin, Stephen and Guo, Baining},
  booktitle={Proceedings of the IEEE/CVF International Conference on Computer Vision (ICCV)},
  year={2021}
}

@inproceedings{clasen2025refinedbigearthnet,
  title={{reBEN}: Refined BigEarthNet Dataset for Remote Sensing Image Analysis},
  author={Clasen, Kai Norman and Hackel, Leonard and Burgert, Tom and Sumbul, Gencer and Demir, Beg{\"u}m and Markl, Volker},
  year={2025},
  booktitle={IEEE International Geoscience and Remote Sensing Symposium (IGARSS)},
}

@article{hong2021SpectralFormer,
  title={Spectralformer: Rethinking hyperspectral image classification with transformers},
  author={Hong, Danfeng and Han, Zhu and Yao, Jing and Gao, Lianru and Zhang, Bing and Plaza, Antonio and Chanussot, Jocelyn},
  journal={IEEE Trans. Geosci. Remote Sens.},
  year={2022},
  volume={60},
  pages={1-15},
  note = {DOI: 10.1109/TGRS.2021.3130716}
}

@article{raulf2020Deep,
author = {Raulf, Arne and Butke, Joshua and Küpper, Claus and Großerüschkamp, Frederik and Gerwert, Klaus and Mosig, Axel},
year = {2019},
month = {06},
pages = {},
title = {Deep representation learning for domain adaptable classification of infrared spectral imaging data},
volume = {36},
journal = {Bioinformatics (Oxford, England)},
doi = {10.1093/bioinformatics/btz505}
}

@Article{he2021SpatialSpectral,
AUTHOR = {He, Xin and Chen, Yushi and Lin, Zhouhan},
TITLE = {Spatial-Spectral Transformer for Hyperspectral Image Classification},
JOURNAL = {Remote Sensing},
VOLUME = {13},
YEAR = {2021},
NUMBER = {3},
ARTICLE-NUMBER = {498},
URL = {https://www.mdpi.com/2072-4292/13/3/498},
ISSN = {2072-4292},
DOI = {10.3390/rs13030498}
}

@article{papoutsis2022Benchmarking,
title = {Benchmarking and scaling of deep learning models for land cover image classification},
journal = {ISPRS Journal of Photogrammetry and Remote Sensing},
volume = {195},
pages = {250-268},
year = {2023},
issn = {0924-2716},
doi = {https://doi.org/10.1016/j.isprsjprs.2022.11.012},
url = {https://www.sciencedirect.com/science/article/pii/S0924271622003057},
author = {Ioannis Papoutsis and Nikolaos Ioannis Bountos and Angelos Zavras and Dimitrios Michail and Christos Tryfonopoulos}
}

@inproceedings{ding2022DaViT,
  title={Davit: Dual attention vision transformers},
  author={Ding, Mingyu and Xiao, Bin and Codella, Noel and Luo, Ping and Wang, Jingdong and Yuan, Lu},
  booktitle={Computer Vision--ECCV 2022: 17th European Conference, Tel Aviv, Israel, October 23--27, 2022, Proceedings, Part XXIV},
  pages={74--92},
  year={2022},
  organization={Springer}
}

@Article{tang2021AdaBoost,
author ="Tang, Jiayi and Henderson, Alex and Gardner, Peter",
title  ="Exploring AdaBoost and Random Forests machine learning approaches for infrared pathology on unbalanced data sets",
journal  ="Analyst",
year  ="2021",
volume  ="146",
issue  ="19",
pages  ="5880-5891",
publisher  ="The Royal Society of Chemistry",
doi  ="10.1039/D0AN02155E",
url  ="http://dx.doi.org/10.1039/D0AN02155E"}

@article{keogan2025Prediction,
author = {Keogan, Abigail and Nguyen, Thi Nguyet Que and Bouzy, Pascaline and Stone, Nicholas and Jirstrom, Karin and Rahman, Arman and Gallagher, William and Meade, Aidan},
year = {2025},
month = {01},
pages = {},
title = {Prediction of post-treatment recurrence in early-stage breast cancer using deep-learning with mid-infrared chemical histopathological imaging},
volume = {9},
journal = {npj Precision Oncology},
doi = {10.1038/s41698-024-00772-x}
}

@misc{tolstikhin2021MLPMixer,
      title={MLP-Mixer: An all-MLP Architecture for Vision}, 
      author={Ilya Tolstikhin and Neil Houlsby and Alexander Kolesnikov and Lucas Beyer and Xiaohua Zhai and Thomas Unterthiner and Jessica Yung and Daniel Keysers and Jakob Uszkoreit and Mario Lucic and Alexey Dosovitskiy},
      year={2021},
      eprint={2105.01601},
      archivePrefix={arXiv},
      primaryClass={cs.CV}
}

@inproceedings{Qin2024MobileNetV4U,
  title={MobileNetV4 - Universal Models for the Mobile Ecosystem},
  author={Danfeng Qin and Chas Leichner and Manolis Delakis and Marco Fornoni and Shixin Luo and Fan Yang and Weijun Wang and Colby R. Banbury and Chengxi Ye and Berkin Akin and Vaibhav Aggarwal and Tenghui Zhu and Daniele Moro and Andrew Howard},
  booktitle={European Conference on Computer Vision},
  year={2024},
  url={https://api.semanticscholar.org/CorpusID:269157379}
}

@inproceedings{he2016resnet,
author = {He, Kaiming and Zhang, Xiangyu and Ren, Shaoqing and Sun, Jian},
year = {2016},
month = {06},
pages = {770-778},
title = {Deep Residual Learning for Image Recognition},
booktitle = {Proceedings of the IEEE Conference on Computer Vision and Pattern Recognition (CVPR)},
doi = {10.1109/CVPR.2016.90}
}

@article {schuhmacher2020CompSegNet,
	author = {Schuhmacher, David and Gerwert, Klaus and Mosig, Axel},
	title = {A Generic Neural Network Approach to Infer Segmenting Classifiers for Disease-Associated Regions in Medical Images},
	elocation-id = {2020.02.27.20028845},
	year = {2020},
	doi = {10.1101/2020.02.27.20028845},
	publisher = {Cold Spring Harbor Laboratory Press},
	URL = {https://www.medrxiv.org/content/early/2020/02/29/2020.02.27.20028845},
	eprint = {https://www.medrxiv.org/content/early/2020/02/29/2020.02.27.20028845.full.pdf},
	journal = {medRxiv}
}

@inbook{woo2018CBAM,
author = {Woo, Sanghyun and Park, JongChan and Lee, Joon-Young and Kweon, Inso},
year = {2018},
month = {09},
pages = {3-19},
title = {CBAM: Convolutional Block Attention Module: 15th European Conference, Munich, Germany, September 8–14, 2018, Proceedings, Part VII},
isbn = {978-3-030-01233-5},
doi = {10.1007/978-3-030-01234-2_1}
}

@INPROCEEDINGS{hu2018SE,
  author={Hu, Jie and Shen, Li and Sun, Gang},
  booktitle={2018 IEEE/CVF Conference on Computer Vision and Pattern Recognition}, 
  title={Squeeze-and-Excitation Networks}, 
  year={2018},
  volume={},
  number={},
  pages={7132-7141},
  doi={10.1109/CVPR.2018.00745}}

@ARTICLE{hang2019cascaded,
  author={Hang, Renlong and Liu, Qingshan and Hong, Danfeng and Ghamisi, Pedram},
  journal={IEEE Transactions on Geoscience and Remote Sensing}, 
  title={Cascaded Recurrent Neural Networks for Hyperspectral Image Classification}, 
  year={2019},
  volume={57},
  number={8},
  pages={5384-5394},
  doi={10.1109/TGRS.2019.2899129}}

@Article{liu2022ConvNeXt,
  author  = {Zhuang Liu and Hanzi Mao and Chao-Yuan Wu and Christoph Feichtenhofer and Trevor Darrell and Saining Xie},
  title   = {A ConvNet for the 2020s},
  journal = {Proceedings of the IEEE/CVF Conference on Computer Vision and Pattern Recognition (CVPR)},
  year    = {2022},
}

@article{Woo2023ConvNeXtV2,
  title={ConvNeXt V2: Co-designing and Scaling ConvNets with Masked Autoencoders},
  author={Sanghyun Woo and Shoubhik Debnath and Ronghang Hu and Xinlei Chen and Zhuang Liu and In So Kweon and Saining Xie},
  year={2023},
  journal={arXiv preprint arXiv:2301.00808},
}

@inproceedings{vaswani2017attention,
 author = {Vaswani, Ashish and Shazeer, Noam and Parmar, Niki and Uszkoreit, Jakob and Jones, Llion and Gomez, Aidan N and Kaiser, \L ukasz and Polosukhin, Illia},
 booktitle = {Advances in Neural Information Processing Systems},
 editor = {I. Guyon and U. Von Luxburg and S. Bengio and H. Wallach and R. Fergus and S. Vishwanathan and R. Garnett},
 pages = {},
 publisher = {Curran Associates, Inc.},
 title = {Attention is All you Need},
 url = {https://proceedings.neurips.cc/paper_files/paper/2017/file/3f5ee243547dee91fbd053c1c4a845aa-Paper.pdf},
 volume = {30},
 year = {2017}
}

@INPROCEEDINGS{wang2021pyramid,
  author={Wang, Wenhai and Xie, Enze and Li, Xiang and Fan, Deng-Ping and Song, Kaitao and Liang, Ding and Lu, Tong and Luo, Ping and Shao, Ling},
  booktitle={2021 IEEE/CVF International Conference on Computer Vision (ICCV)}, 
  title={Pyramid Vision Transformer: A Versatile Backbone for Dense Prediction without Convolutions}, 
  year={2021},
  volume={},
  number={},
  pages={548-558},
  doi={10.1109/ICCV48922.2021.00061}}

@ARTICLE{9325094,
  author={Xue, Zhaohui and Zhang, Mengxue and Liu, Yifeng and Du, Peijun},
  journal={IEEE Transactions on Geoscience and Remote Sensing}, 
  title={Attention-Based Second-Order Pooling Network for Hyperspectral Image Classification}, 
  year={2021},
  volume={59},
  number={11},
  pages={9600-9615},
  doi={10.1109/TGRS.2020.3048128}}

@Article{s23052499,
AUTHOR = {Uchaev, Denis and Uchaev, Dmitry},
TITLE = {Small Sample Hyperspectral Image Classification Based on the Random Patches Network and Recursive Filtering},
JOURNAL = {Sensors},
VOLUME = {23},
YEAR = {2023},
NUMBER = {5},
ARTICLE-NUMBER = {2499},
URL = {https://www.mdpi.com/1424-8220/23/5/2499},
PubMedID = {36904702},
ISSN = {1424-8220},
DOI = {10.3390/s23052499}
}

@inproceedings{tishby2000information,
author = {Tishby, Naftali and Pereira, Fernando and Bialek, William},
year = {2001},
month = {07},
pages = {},
title = {The Information Bottleneck Method},
volume = {49},
booktitle = {Proceedings of the 37th Allerton Conference on Communication, Control and Computation},
doi = {10.48550/arXiv.physics/0004057}
}

@inproceedings{hjelm2018learning,
title={Learning deep representations by mutual information estimation and maximization},
author={R Devon Hjelm and Alex Fedorov and Samuel Lavoie-Marchildon and Karan Grewal and Phil Bachman and Adam Trischler and Yoshua Bengio},
booktitle={International Conference on Learning Representations},
year={2019},
url={https://openreview.net/forum?id=Bklr3j0cKX},
}

@inproceedings{higgins2017beta,
title={beta-{VAE}: Learning Basic Visual Concepts with a Constrained Variational Framework},
author={Irina Higgins and Loic Matthey and Arka Pal and Christopher Burgess and Xavier Glorot and Matthew Botvinick and Shakir Mohamed and Alexander Lerchner},
booktitle={International Conference on Learning Representations},
year={2017},
url={https://openreview.net/forum?id=Sy2fzU9gl}
}

@inproceedings{locatello2019challenging,
  title={Challenging Common Assumptions in the Unsupervised Learning of Disentangled Representations},
  author={Francesco Locatello and Stefan Bauer and Mario Lucic and Sylvain Gelly and Bernhard Scholkopf and Olivier Bachem},
  booktitle={International Conference on Machine Learning},
  year={2018},
  url={https://api.semanticscholar.org/CorpusID:54089884}
}

@inproceedings{wang2020eca,
   title={ECA-Net: Efficient Channel Attention for Deep Convolutional Neural Networks},
   author={Qilong Wang and Banggu Wu and Pengfei Zhu and Peihua Li and Wangmeng Zuo and Qinghua Hu},
   booktitle = {The IEEE Conference on Computer Vision and Pattern Recognition (CVPR)},
   year={2020}
 }

@article{cao2019gcnet,
  title={GCNet: Non-Local Networks Meet Squeeze-Excitation Networks and Beyond},
  author={Yue Cao and Jiarui Xu and Stephen Lin and Fangyun Wei and Han Hu},
  journal={2019 IEEE/CVF International Conference on Computer Vision Workshop (ICCVW)},
  year={2019},
  pages={1971-1980},
  url={https://api.semanticscholar.org/CorpusID:131775495}
}

@inproceedings{li2019selective,
  title={Selective Kernel Networks},
  author={Li, Xiang and Wang, Wenhai and Hu, Xiaolin and Yang, Jian},
  booktitle={IEEE Conference on Computer Vision and Pattern Recognition},
  year={2019}
}

@INPROCEEDINGS{szegedy2015going,
  author={Szegedy, Christian and Wei Liu and Yangqing Jia and Sermanet, Pierre and Reed, Scott and Anguelov, Dragomir and Erhan, Dumitru and Vanhoucke, Vincent and Rabinovich, Andrew},
  booktitle={2015 IEEE Conference on Computer Vision and Pattern Recognition (CVPR)}, 
  title={Going deeper with convolutions}, 
  year={2015},
  volume={},
  number={},
  pages={1-9},
  doi={10.1109/CVPR.2015.7298594}}

@InProceedings{graham2021levit,
    author    = {Graham, Benjamin and El-Nouby, Alaaeldin and Touvron, Hugo and Stock, Pierre and Joulin, Armand and Jegou, Herve and Douze, Matthijs},
    title     = {LeViT: A Vision Transformer in ConvNet's Clothing for Faster Inference},
    booktitle = {Proceedings of the IEEE/CVF International Conference on Computer Vision (ICCV)},
    month     = {October},
    year      = {2021},
    pages     = {12259-12269}
}

@misc{dong2022cswin,
      title={CSWin Transformer: A General Vision Transformer Backbone with Cross-Shaped Windows}, 
        author={Xiaoyi Dong and Jianmin Bao and Dongdong Chen and Weiming Zhang and Nenghai Yu and Lu Yuan and Dong Chen and Baining Guo},
        year={2021},
        eprint={2107.00652},
        archivePrefix={arXiv},
        primaryClass={cs.CV}
}

@article{tu2022maxvit,
  title={MaxViT: Multi-Axis Vision Transformer},
  author={Tu, Zhengzhong and Talebi, Hossein and Zhang, Han and Yang, Feng and Milanfar, Peyman and Bovik, Alan and Li, Yinxiao},
  journal={ECCV},
  year={2022},
}

@inproceedings{park2018bam,
  title={BAM: Bottleneck Attention Module},
  author={Jongchan Park and Sanghyun Woo and Joon-Young Lee and In-So Kweon},
  booktitle={British Machine Vision Conference},
  year={2018},
  url={https://api.semanticscholar.org/CorpusID:49864419}
}

@inproceedings{chen2017deeplab,
  title={Rethinking atrous convolution for semantic image segmentation},
  author={Chen, Liang-Chieh and Papandreou, George and Schroff, Florian and Adam, Hartwig},
  booktitle={arXiv preprint arXiv:1706.05587},
  year={2017}
}

@misc{PURR1947,
	title = {220 Band AVIRIS Hyperspectral Image Data Set: June 12, 1992 Indian Pine Test Site 3},
	month = {Sep},
	url = {https://purr.purdue.edu/publications/1947/1},
	year = {2015},
	doi = {doi:/10.4231/R7RX991C},
	author = {Marion F. Baumgardner and and Larry L. Biehl and David A. Landgrebe }
}

@misc{datasetpavia,
  author = {Gamba, Paolo and Dell'Acqua, Fabio},
  title  = {Pavia University ROSIS Hyperspectral Data},
  year   = {2002},
  howpublished = {University of Pavia (via GIC, EHU)},
  note   = {Available: \url{https://www.ehu.eus/ccwintco/index.php/Hyperspectral_Remote_Sensing_Scenes}}
}

@article{houston2013,
  author   = {Debes, Christian and Merentitis, Andreas and Heremans, Roel and Hahn, J{\"u}rgen and Frangiadakis, Nikolaos and van Kasteren, Tim and Liao, Wenzhi and Bellens, Rik and Pi{\v{z}}urica, Aleksandra and Gautama, Sidharta and Philips, Wilfried and Prasad, Saurabh and Du, Qian and Pacifici, Fabio},
  title    = {Hyperspectral and {LiDAR} Data Fusion: Outcome of the 2013 {GRSS} Data Fusion Contest},
  journal  = {IEEE Journal of Selected Topics in Applied Earth Observations and Remote Sensing},
  year     = {2014},
  volume   = {7},
  number   = {6},
  pages    = {2405-2418},
  doi      = {10.1109/JSTARS.2014.2305441}
}
}

\end{document}